\documentclass[letterpaper, 10 pt, conference]{ieeeconf}  

\IEEEoverridecommandlockouts                              

\usepackage{graphicx}
\usepackage{amsmath,amssymb} 
\usepackage{color}
\usepackage{caption}
\usepackage{subcaption}
\usepackage{makecell}
\usepackage{multirow}

\newcommand{\env}{WorldViewEnv}
\newcommand{\name}{NavNet}

\title{\LARGE \bf
Long Range Neural Navigation Policies for the Real World}

\author{Ayzaan Wahid$^{1}$, Alexander Toshev$^{1}$, Marek Fiser$^{1}$ and Tsang-Wei Edward Lee$^{1}$
\thanks{Authors are with Google AI, 1600 Amphitheatre Pkwy, Mountain View, CA 94043, USA. Corresponding author Alexander Toshev, {\tt\small toshev@google.com}}%
}

\begin{document}

\maketitle
\thispagestyle{empty}
\pagestyle{empty}


\begin{abstract}
Learned Neural Network based policies have shown promising results for robot navigation. However, most of these approaches fall short of being used on a real robot due to the extensive simulated training they require. These simulations lack the visuals and dynamics of the real world, which makes it infeasible to deploy on a real robot. We present a novel Neural Net based policy, \name, which allows for easy deployment on a real robot. It consists of two sub policies -- a high level policy which can understand real images and perform long range planning expressed in high level commands; a low level policy that can translate the long range plan into low level commands on a specific platform in a safe and robust manner. For every new deployment, the high level policy is trained on an easily obtainable scan of the environment modeling its visuals and layout. We detail the design of such an environment and how one can use it for training a final navigation policy. Further, we demonstrate a learned low-level policy. We deploy the model in a large office building and test it extensively, achieving $0.80$ success rate over long navigation runs and outperforming SLAM-based models in the same settings.
\end{abstract}

\section{Introduction}\label{sec:intro}
Robot navigation is one of the fundamental challenges in robotics needed for autonomous intelligent agents. This problem is traditionally defined as finding a path from a start location to a target location and executing this path in a robust and safe manner, e.g.\ go from the office kitchen to the whiteboard. It requires the ability for the robot to perceive its environment, localize itself w.r.t.\ a target, reason about obstacles in its immediate vicinity, and develop a long range plan for getting to the target.

There is a huge body of work on navigation~\cite{bonin2008visual}. Traditionally, navigation systems rely on feature extraction and geometric based reasoning to localize a robot and map its surroundings~\cite{thrun2007simultaneous}. When such maps are generated or given, the robot can use them to find a navigation path using planning algorithms~\cite{thrun2005probabilistic}.

More recently, learning navigation policies has emerged as a new line of work. Learned policies have the potential to leverage all aspects of the observations relevant to navigation, and not only the ones encoded manually in a geometric map of the environment. Furthermore, they can learn navigation behaviors which are not expressable with traditional planners.

Most of the learned policies capitalize on the recent advances in Deep Reinforcement Learning (RL). Despite their potential, the current approaches have been difficult to successfully deploy on real robots. Due to the high sample complexity of RL algorithms, these neural policies can often only be successfully trained in simulation environments. For navigation problems, such environments ought to capture both the physics and the visual complexity of the real world. Having both in one simulation environment is very challenging and practically non-existent (see Sec.~\ref{sec:related}).

\begin{figure}[t]
\centering
\includegraphics[width=0.97\columnwidth]{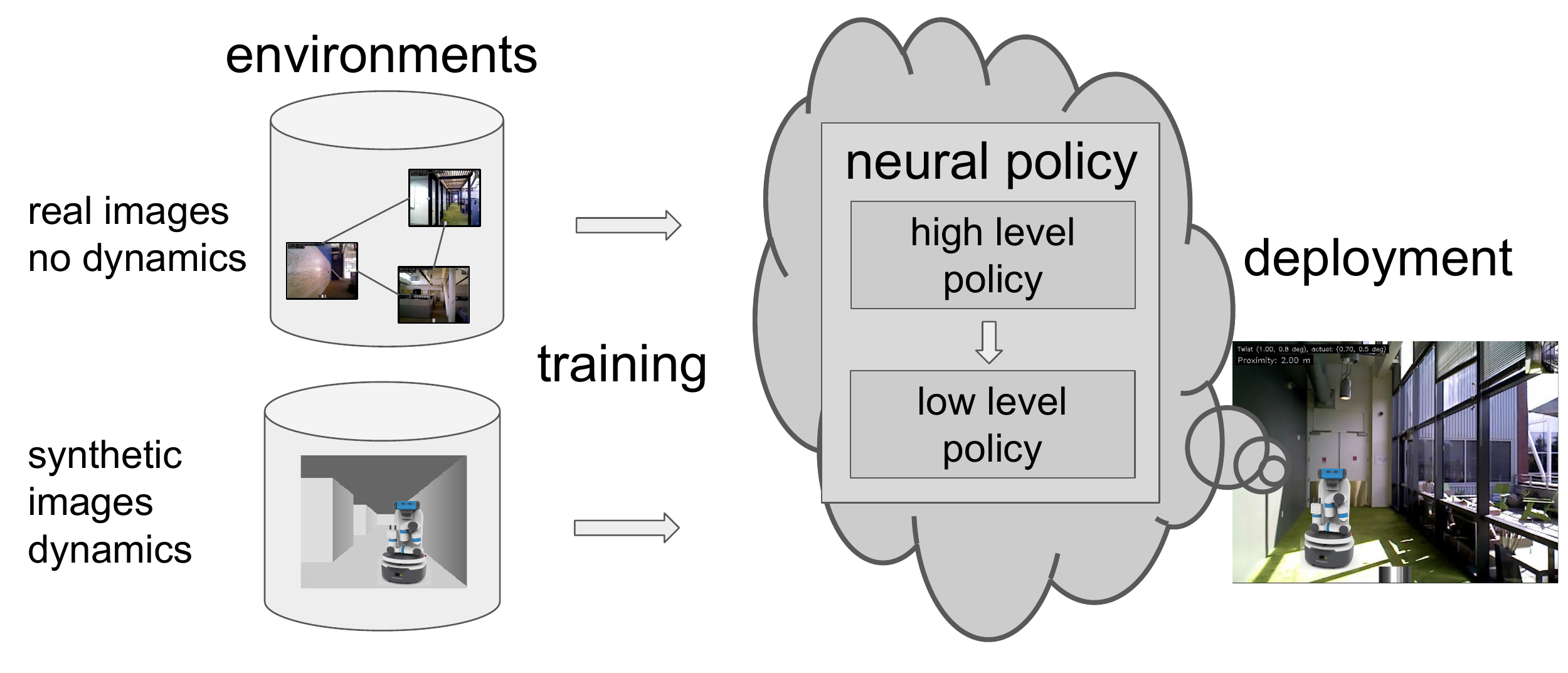}
\caption{\label{fig:intro}Use of two separate environments, the first modeling the visuals and layout of the deployment environment, the second modeling the robot physics, allows for easy training and deployment of neural policies in the real world.}
\end{figure}

\textbf{Contributions} In this work, we present a neural net based policy for navigation, \name, which is (i) fully learned and (ii) directly deployable on a real robot (see Fig.~\ref{fig:intro}). This is in contrast to most of learned policies which are not directly usable on a real robot (see Sec.~\ref{sec:related}). Our policy is a two level policy -- long range and short range. The long range is responsible for high level commands such as 'go forward', 'turn left', etc. while the short range policy executes those commands. The high level policy is trained to interpret real images from the deployment environment for the purpose of planning. While technically any local planner can be used to execute commands given by the high level policy, we show that training a short range policy specifically to complement the needs of the high level planner on a mobile base -- being precise, keeping a straight path, and avoiding collisions -- results in better performance in the real world.

This hierarchical separation into two policies, with the specified interface between them, allows for training these on different environments. As a second contribution, we propose to train the high-level on a 'streetview'-like environment of our indoor spaces. Training environments which exhibit high visual truthfulness while precisely modeling robot physics are hard to create. However, images of the deployment space organized in a semi-metric graph are easy to obtain. We show how to easily construct such a graph and train a high-level policy on it. Furthermore, simple maze-like synthetic environments, which model a robot mobile base, can be built using simulators. We train a low-level policy on such a synthetic environment.

We apply and extend \cite{mousavian2018visual} to learn our high-level policy. The supervision can come from a traditional path planner, as done in \cite{mousavian2018visual}, or from human labeled paths. As a further contribution, we show how to utilize human demonstrations of navigation paths. These demonstrations allow for the robot to learn behaviors not easily expressed by traditional planners.

We extensively test the presented approach in a large ($200m x 100m$) office building. We demonstrate that we can achieve $0.80$ success rate navigating to $10$ targets from random starting points across a building on a real robot, which is higher than the performance of a SLAM-based system in the same environment. Furthermore, our performance is quantitatively new from published neural models, which have not been deployed in the real world at this scale.

\section{Related Work}\label{sec:related}
\textbf{SLAM-based Navigation} As a very old problem, there is an extensive literature on SLAM, planning and robot navigation~\cite{desouza2002vision,bonin2008visual,thrun2007simultaneous,armeni_cvpr16}, which due to space limitations will not be discussed here. In a broad sense, our work falls into mapless navigation. The neural policy requires neither a geometric map of the environment nor localization at test time.

\textbf{Neural Net Based Navigation} 
In recent years, RL-learned neural net policies have been explored for navigation. These are usually learned and tested in simulation. Examples include: using A3C on 3D mazes \cite{mirowski2016learning}; A3C on AITHOR~\cite{zhu2017target}; ADDPG tested using a depth sensor on a real robot~\cite{tai2017virtual}; RL algorithms trained and tested on scans only of real spaces (no real robot)~\cite{savva2017minos,gupta2017cognitive} or SUNCG~\cite{wu2018building}. Due to the large sample complexity of RL, the above methods cannot be learned directly on a real robot.

A high level environment has been used by Mirowski et al.~\cite{mirowski2018learning} without deployment on a real robot. Bruce et al.~\cite{bruce2018learning,bruce2017one} deploy a RL trained neural net policy and use an environment constructed from a traversal. However, their system is not fully deployed on a real robot -- the policy actions are executed by an operator, while our system employs a second low-level policy to execute these actions. Thus, contrary to them we provide a fully deployable navigation solution.

An explicit path planning strategy, similar to our approach, has been employed by Savinov et al.~\cite{savinov2018semi}, in a learned topological graph. However, this path planning is used for inference only, and the system is evaluated in synthetic environments.

Beyond RL algorithms, investigations have been conducted into appropriate architectures, with emphasis on using models with external memory~\cite{zhangneural,parisotto2017neural,oh2016control,khan2017memory}. These approaches, in their current form, are only applied in simulation. Learned low level controllers have been developed~\cite{chiang2018learning}, and also combined with traditional PRM planning~\cite{faust2017prm}.

\textbf{Realistic Training Environments for Navigation} The majority of the simulation environments used for navigation experiments are not visually realistic, and as such the trained policies are not usable on a robot in the real world. Some of these environments use photo realistic rendering \cite{dosovitskiy2017carla,zhu2017target}, but are still not extensively tested in real settings.

Recently, captures of real world environments are available in the form of train/test environments for robotics. The closest match to our desired environment is the Active Vision Dataset~\cite{ammirato2017dataset}, which is a dense capture of homes. Unfortunately, these environments, while large in number, are individually too small to sufficiently challenge long range navigation. Larger environments (\cite{armeni2017joint, chang2017matterport3d, xia2018gibson}), are captured too sparsely and contain rendered 3D reconstructions for locations not in the original scan. Although the reconstructions are of high visual fidelity, the rendered views still contain artifacts.

\section{Navigation Model}\label{sec:model}

\begin{figure}[t]
\includegraphics[width=0.9\columnwidth]{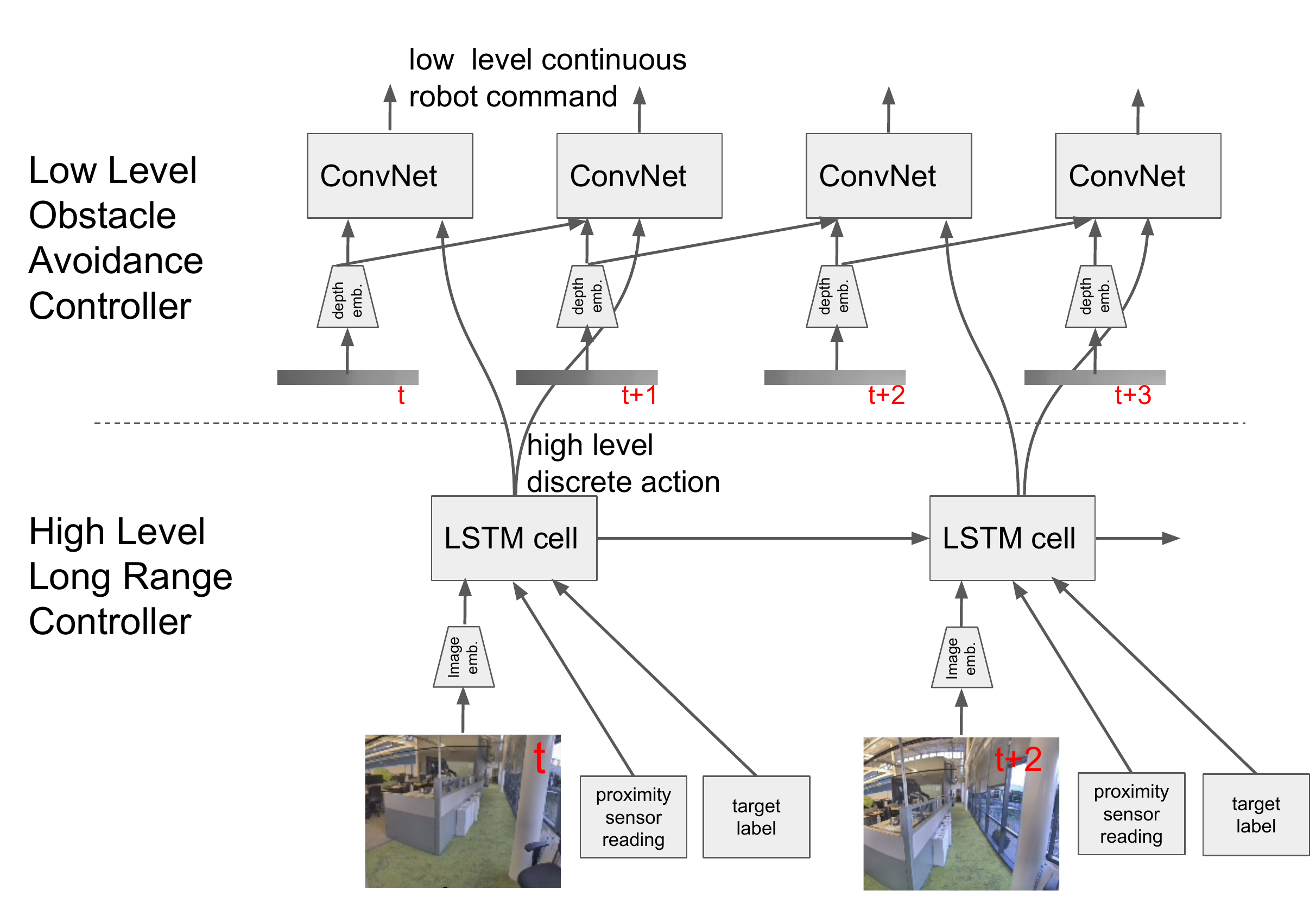}
\caption{\label{fig:model}Two level navigation policy. Bottom: the high level navigation policy uses rich RGB data to perform long range planning using a simple high level action space. Top: the high level actions are executed by a low level policy, which perceives depth and avoids obstacles. See Sec.~\ref{sec:model}.}
\end{figure}

The navigation model is a two level policy (see Fig.~\ref{fig:model}). The high level policy is responsible for long range planning -- for every observation and target label, it is trained to output a high level action bringing the robot closer to the target. The low level policy is responsible for executing the high level action on the specified platform. It is trained to maintain a straight path and safe distance to nearby objects.

The two policies have complementary properties. The high level policy knows and plans in the deployment environment, but cannot precisely guide the robot. The low level policy has no knowledge of the deployment environment, as it is trained on a different simulation environment, but moves the robot precisely and safely (see Table~\ref{tab:comparison}).

\begin{table}[b]
\centering
\begin{tabular}{|l|l|l|}
\hline
& \textbf{High Level Planning} & \textbf{Low Level Control} \\
\hline\hline
\multicolumn{3}{|c|}{\textit{Policy Properties}} \\
\hline
inputs & RGB images & 1D depth \\
\hline
actions & \makecell[l]{discrete / high level \\ navigation directions} & \makecell[l]{continuous / twist \\ for diff.~drive} \\
\hline\hline 
\multicolumn{3}{|c|}{\textit{Training Environment Simulation Quality}} \\
\hline
\makecell[l]{sensor  readings} & high & medium \\
\hline
\makecell[l]{robot physics} & none & high \\
\hline
\makecell[l]{building layout} & yes & no \\
\hline
\end{tabular}
\caption{\label{tab:comparison} Comparison of the two policies and their training environments. See Sec.~\ref{sec:model} and \ref{sec:training_environment}.}
\end{table}

\subsection{Long Range High Level Planning Policy}\label{sec:long_range_policy}
The high level policy takes as inputs an RGB image $x$, a target specification $g$, and a binary proximity indicator $p\in\{0,1\}$. The latter indicator is the output of a radar reading and indicates if a collision is imminent. Currently, it is defined as $1$ iff there is an object within $0.3$ m.

We use two ways of specifying targets: a location label or a target image. To define a location label, we overlay a regular grid onto the world and define each cell as a potential target. In our case, we have a $10\times 10$ grid where $77$ cells represent valid locations. Each location is defined as a one hot vector $g\in\{0,1\}^k$, where $k$ is the number of all possible locations.

The image-based target specification is defined by an embedding $g\in\mathbb{R}^d$ of an image from the target location. The embedding is obtained with same network used to embed observations. This second target definition is more flexible, but also more ambiguous as locations might look quite similar in visually repetitive environments. 

The output action space is defined as three possible actions $\mathcal{A}_{\textrm{high}}=\{\textrm{`forward'}, \textrm{`turn left'}, \textrm{`turn right'}\}$. The forward motion is intended to be $1m$, the turns are at $15^\circ$. Note, however, that these values are approximate, as their semantics are established during training of the policy. The training environment, however, does not represent the above actions with very high precision (see Sec.~\ref{sec:high_environment})

With the above notation, the high level policy is trained to output a value $v(a, x; g)$ estimating the progress towards the target $g$, defined as the negative change in distance to $g$ if action $a$ is taken at observation $x$. This value function can be used to estimate which action moves the robot closest to the target:
\begin{equation}
    a_{\textrm{high}} = \arg\max_{a\in\mathcal{A}_{\textrm{high}}}v(a, x; g)
\end{equation}

The above value function is implemented as a recurrent neural net taking as input the concatenated and transformed embeddings of the observation $x$, target $g$, and the proximity bit $p$:
\begin{equation}\label{eq:high_level_architecture}
    v(a, x; g) = LSTM(\textrm{MLP}_2(\textrm{ResNet50}(x), p, \textrm{MLP}_1(g)))
\end{equation}

The recurrent net is a single layer LSTM~\cite{hochreiter1997long}, which maintains a state over the execution of a navigation run. This state serves to capture recent actions, which has shown cricual in avoiding oscillations (revisiting the same state over and over again).

Before being fed into the LSTM, the observation is embedded using an ImageNet pre-trained ResNet50~\cite{he2016deep}. The target embedding is obtained using a single layer perceptron from the location label or image-based target specification. The dimensions of the above perceptrons and LSTM are set to $2048$.

\subsection{Low-Level Policy}
While the high level commands can be executed on the robot using position control, their verbatim execution leads to inaccuracies, drifts, non-smooth behavior and eventually collisions. With the above definition of $\mathcal{A}_{\textrm{high}}$, we can safely execute rotations, but for 'forward' we use a short range policy. This policy is trained to rectify inaccuracies of the high level plan, keep a straight path and avoid collisions.

The input to the policy is a 1-dimensional LiDAR reading. It captures the geometry of nearby objects, which is sufficient for short term safe control. The low level action space $\mathcal{A}_{\textrm{low}}$ is continuous, defined by the kinematics of the robot. In our case, we work with a differential drive mobile base. Thus, the action space is a 2 dimensional real valued vector of the twist values for the two wheels (angular velocities for each wheel). 

The policy is formulated as a ConvNet over both space and time. Since it is a ConvNet, it maintains no state. However, having as an input several recent observations allows for the policy to reason over robot motion, which we believe is important for low-level control.

In more details, the low level policy receives the last 3 LiDAR readings $x$, $x_{-1}$ and $x_{-2}$. Since these are 1-dimensional, they can be concatenated into an image, where the second dimension is time:
\begin{equation}\label{eq:low_level}
    a_{\textrm{low}} = ConvNet(\textrm{concat}(x_{-2}, x_{-1}, x))
\end{equation}
where $a_{\textrm{low}}\in\mathcal{A}_{\textrm{low}}$. 

The network $ConvNet$ has following 4 layers: $conv([7, 3, 16], 5)\rightarrow conv([5, 1, 20], 3)\rightarrow fc(20) \rightarrow fc(10)$, where $conv(k, s)$ denotes convolution with kernel $k$ and stride $s$ and $fc(d)$ is fully connected layer with output dimension $d$.

\section{Training Environments and Algorithms}\label{sec:training_environment}
\subsection{High Level Planning Policy}\label{sec:high_environment}

\textbf{\env} The training uses an environment consisting of real images $\mathcal{X}$ obtained via traversals. These images represent states of the robot in the world and are organized in a graph, whose edges represent actions moving the robot from one state to another. 

\begin{figure}[t]
\centering
\includegraphics[width=0.7\columnwidth]{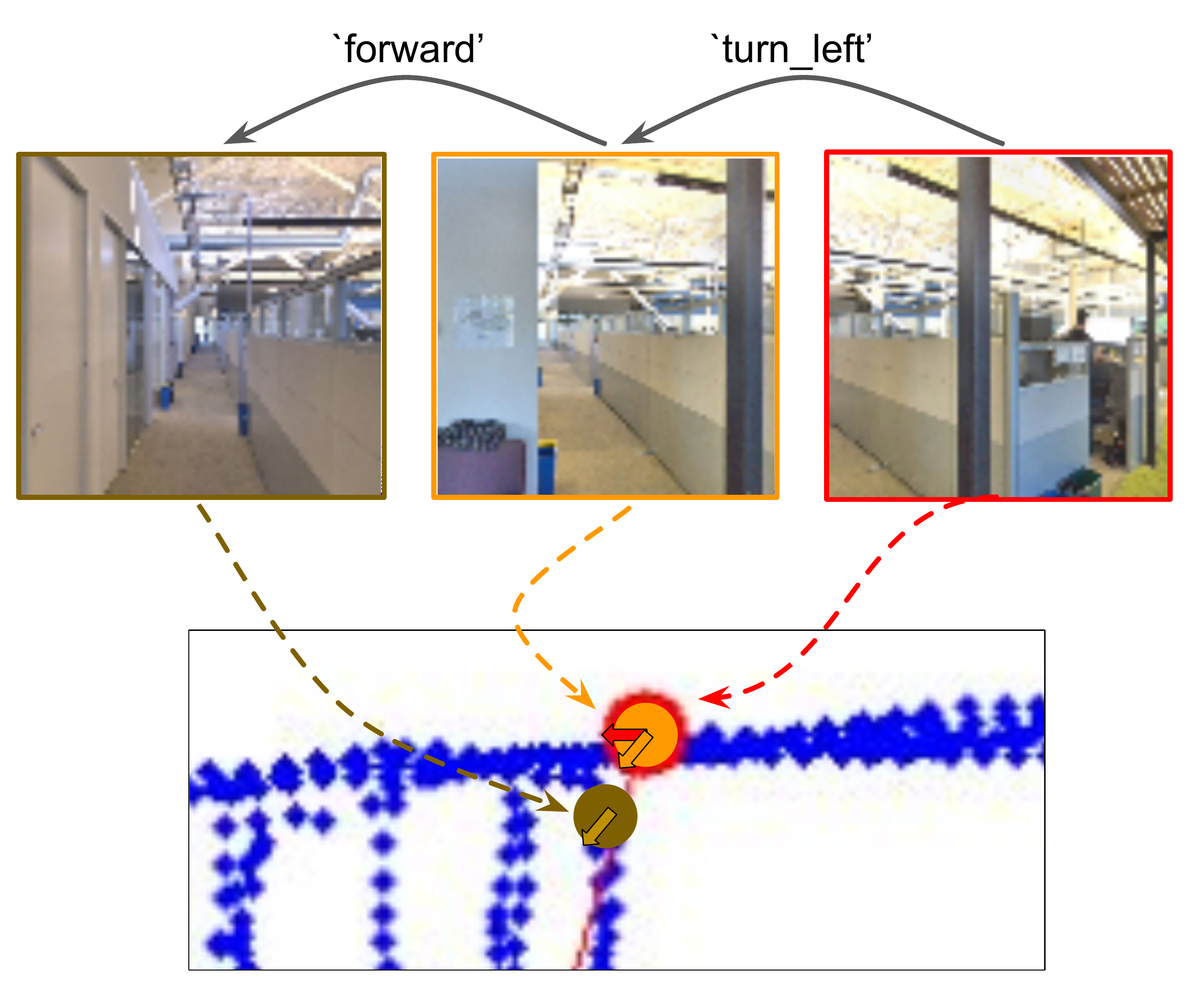}
\caption{\label{fig:worldviewenv-traversal}An example 2 step traversal of \env{}. Top: views at three different states. Bottom: Top down view of all traversed locations in the vicinity of the views (as blue dots). The enlarged circles correspond to the three views, the arrow represents the robot orientation.
}
\end{figure}

To create \env{}, we use a rig of 6 cameras organized in a hexagonal shape. This rig captures a set of images every $0.5m$ as it moves through the building.

We perform two operations with the raw captured data. First, the images are stitched into a 360 degree panorama, which can be cropped in any direction to obtain images of the desired field of view (fov). We use fov of 108 deg (width) and 90 deg (height). We crop each panorama into 24 views each 15 degrees apart. We define two rotational actions: `turn left' and `turn right', which shifts the robot's view while remaining at the same location.

Second, we use the Cartographer API~\cite{quigley2009ros} to estimate the poses of the images. We define the `forward' action as moving the robot to a nearby location in the direction of the current view. We attempt to move forward by $1m$, but there is no guarantee that an image was captured exactly $1m$ away. We consider this action possible if an image is captured within $0.7m$ from the new location, which we set as the result of this action. The final action space is $\mathcal{A}_{\textrm{high}}=\{\textrm{`forward'}, \textrm{`turn left'}, \textrm{`turn right'}\}$ (see Fig.~\ref{fig:worldviewenv-traversal}).

\textbf{Discussion} \env{} exhibits high visual fidelity and the  traversals cover most of the designated building spaces. However, the high level actions capture rough motions and can be used to express a navigation path, but cannot be executed robustly on a robot. Hence, we use this environment to train a high level policy only.

As the actions connecting views are high level, we need only locally correct SLAM and loop closure. Hence, we do not need the high precision necessary for global geometric maps. Moreover, no mapping of the surroundings is needed.

\textbf{Training} The training is a form of imitation learning, where we can imitate paths from traditional planners and human navigations. Furthermore, imitation learning provides supervision at every step of the policy execution, which leads to more stable optimization contrary to many of the RL setups suffering from sparse rewards.

These paths can be used as supervision at every step of the policy execution, when present. Supervised learning is more efficient than RL due to the lower sample complexity.  

To define the training loss, consider a set of navigation paths $\mathcal{P}=\{p_1, ..., p_N\}$ leading to one of several pre-defined targets. These paths are defined over the graph underlying \env{}. In our experiments $\mathcal{P}$ is either a set of all shortest paths to targets produced by a shortest path planner or a set of human navigation paths.

For a target $g$, start $x$ and a path $p\in\mathcal{P}$, we can denote by $d(x, g; p)$ the distance from $x$ to $g$ along $p$ if both start and target are on the path in this order. If one of or both of them are not on the path, then the above distance is $\inf$. Using the full $\mathcal{P}$, we can consider the shortest path in $\mathcal{P}$ which leads from $x$ to $g$:
\begin{equation*}
    d(x, g;\mathcal{P}) = \min_{p\in\mathcal{P}}d(x, g; p)
\end{equation*}

Using $d$, we can define the progress toward $g$ if we apply $a$ at state $x$:
\begin{equation*}
    y(a, x; g) = d(x, g; \mathcal{P}) - d(x', g; \mathcal{P})
\end{equation*}
where $x'$ is the image at which one arrives after taking action $x$. This transition is defined by the \env{}.

The loss trains the high level policy to output values as close a possible to $y$. As we use a recurrent net to define the policy (see Eq.~(\ref{eq:high_level_architecture})), we define the loss over whole navigation paths. Denote by $\mathbf{x}=(x_1, \ldots, x_T)$ a navigation path, then the loss reads
\begin{equation*}
    Loss(\textbf{x},g) = \sum_{x_t\in\mathbf{x}}\sum_{a\in\mathcal{A}_{\textrm{high}}} (v(a, x_t; g) - y(a, x_t; g))^2
\end{equation*}
where the model $v$ is defined in Eq.~(\ref{eq:high_level_architecture}). We use Adam Optimizer~\cite{kingma2014adam}, where at each training step we unroll the current policy with random starts and formulate the above loss for states which are covered by $\mathcal{P}$. At the beginning of the training, the policy performs random actions, resulting in random paths. As the training progresses, the paths become more meaningful and the above loss emphasizes on situations which will be encountered at test time. This approach is very similar to DAGGER~\cite{ross2011reduction}.

We utilize a scalable distributed training setup~\cite{horgan2018distributed} in which the unrolling is done in separate processes for performance reasons. More precisely, we run $200$ collector jobs, which refresh their policy every $10K$ steps. The unrolled episodes are written into a replay buffer. $5$ training jobs utilize the data in the buffer to perform the actual training. We use standard data augmentation techniques for images, learning rate of $0.0001$, batch size of $8$, and unrolls of length of $40$.

\subsection{Low Level Policy}\label{sec:low_environment}
\begin{figure}[t]
\centering
\includegraphics[width=0.7\columnwidth]{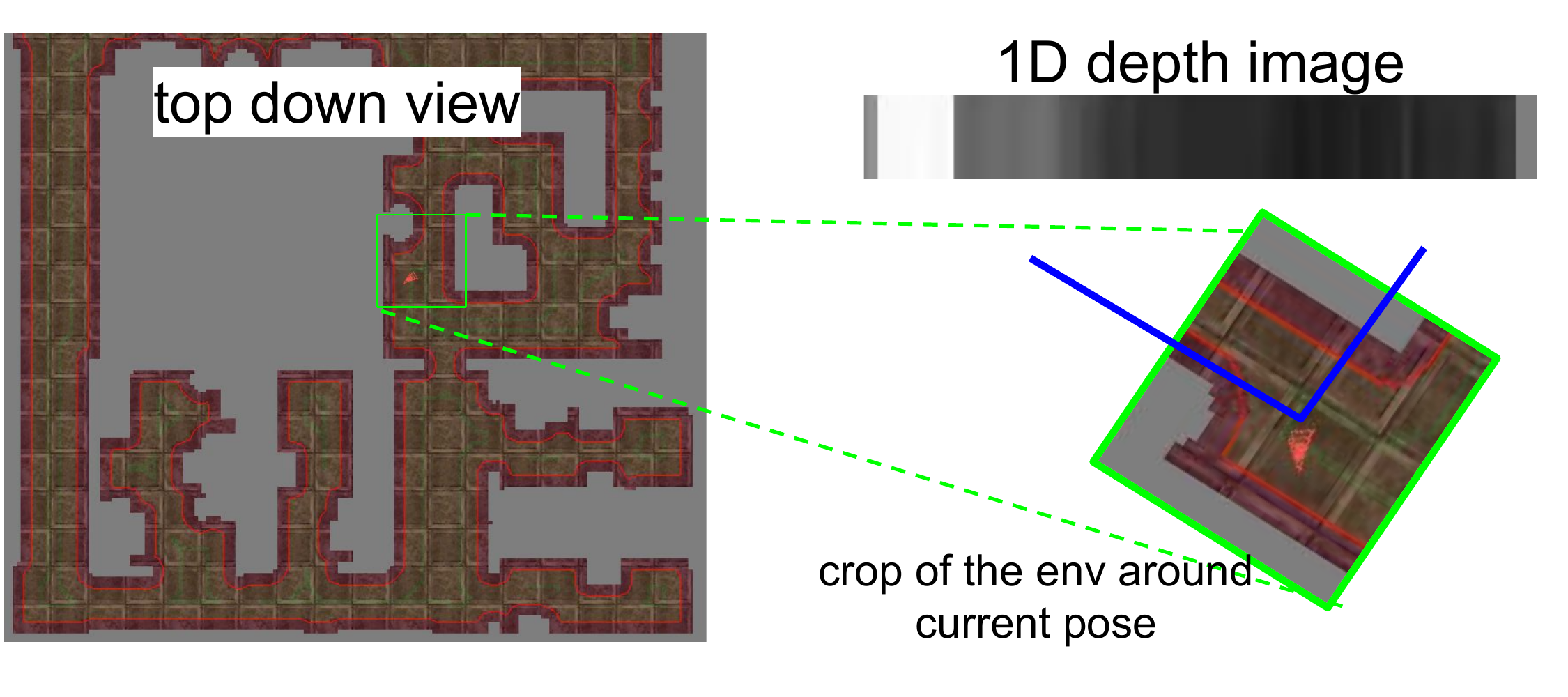}
\caption{\label{fig:low-level-env}Left: top down view of the environment. Brown areas are traversable. Right: visualization of the 1D depth image, which is returned as an observation at the given location and orientation. The blue lines show the field of view.}
\end{figure}

\textbf{Simple Synthetic Environment}
The low level policy is trained in a synthetic environment, consisting of several hallways and rooms (see Fig.~\ref{fig:low-level-env}, left). It is generated using a 2D layout, which is converted to 3D by extending the walls up. The observations are 1D depth images (see Fig.~\ref{fig:low-level-env}, right). Due to their simplicity, these can be simulated with high fidelity and the trained models transfer to the real robot. Visually, this environment is a maze of hallways of varying sizes. 

In addition, we simulate the kinematics of the robot, a differential drive. It consists of two wheels which are controlled by their velocities. We do not simulate the robot's dynamics. 

Thus, we simulate the robot motion with simple depth perception. The experiments show that this is sufficient to learn a low level policy.

\textbf{Training}
Our policy is to execute the `forward' action while being efficient and safe. For training we use continuous Deep Q-Learning (DDPG) \cite{lillicrap2015continuous}. The above requirements are encoded by the reward $R(x, a)$ needed by DDPG for a given action $a$ at a state $x$. The reward should be the highest if the robot is moving straight as quickly as possible without colliding:
\begin{eqnarray*}
R(x, a) = \left\{
\begin{array}{ll}R_{lin}v_{lin}(a) + R_{ang}|v_{ang}(a)|& \quad\textrm{no collision} \\
R_{collision}&\quad\textrm{collision}
\end{array}\right.
\end{eqnarray*}
where $v_{lin}(a)$ and $v_{ang}(a)$ denote the linear and angular velocity of the differential drive after applying the current action $a$ (in the current state which is omitted for brevity). If this action results in no collision, the reward is a function of how fast ($R_{lin} = 1.0$) and how straight ($R_{ang} = -0.8$) the robot moves. If there is a collision, then the robot incurs a large negative reward $R_{collision} = -1.0$.

The above setup results in policies which are smooth and maintain a large distance from their surroundings. Empirically, they tend to follow the medial axis of hallways and rooms.

\textbf{Implementation} We borrow the implementation details and the training setup of ~\cite{chiang2018learning}, where we replace their reward with the one above. To outline, we use Adam optimizer~\cite{kingma2014adam} with learning rate of $0.0001$ and batch size of $256$. The employed DDPG algorithm uses a `critic network', which approximates the Q value for given state $x$ and action $a$. It has a very similar architecture to the low level policy, defined in Eq.~\ref{eq:low_level} -- it uses the same type of ConvNet to embed the observation, which is subsequently concatenated with the action. This vector is embedded to a single value using a two layer perceptron (layers of dimensions 10 and 8) .

\section{Experiments}

To assess performance, we measure \textit{success rate} as the portion of the runs which end within $3m$ of the specified target.

\begin{figure}
\centering
\includegraphics[width=0.65\columnwidth]{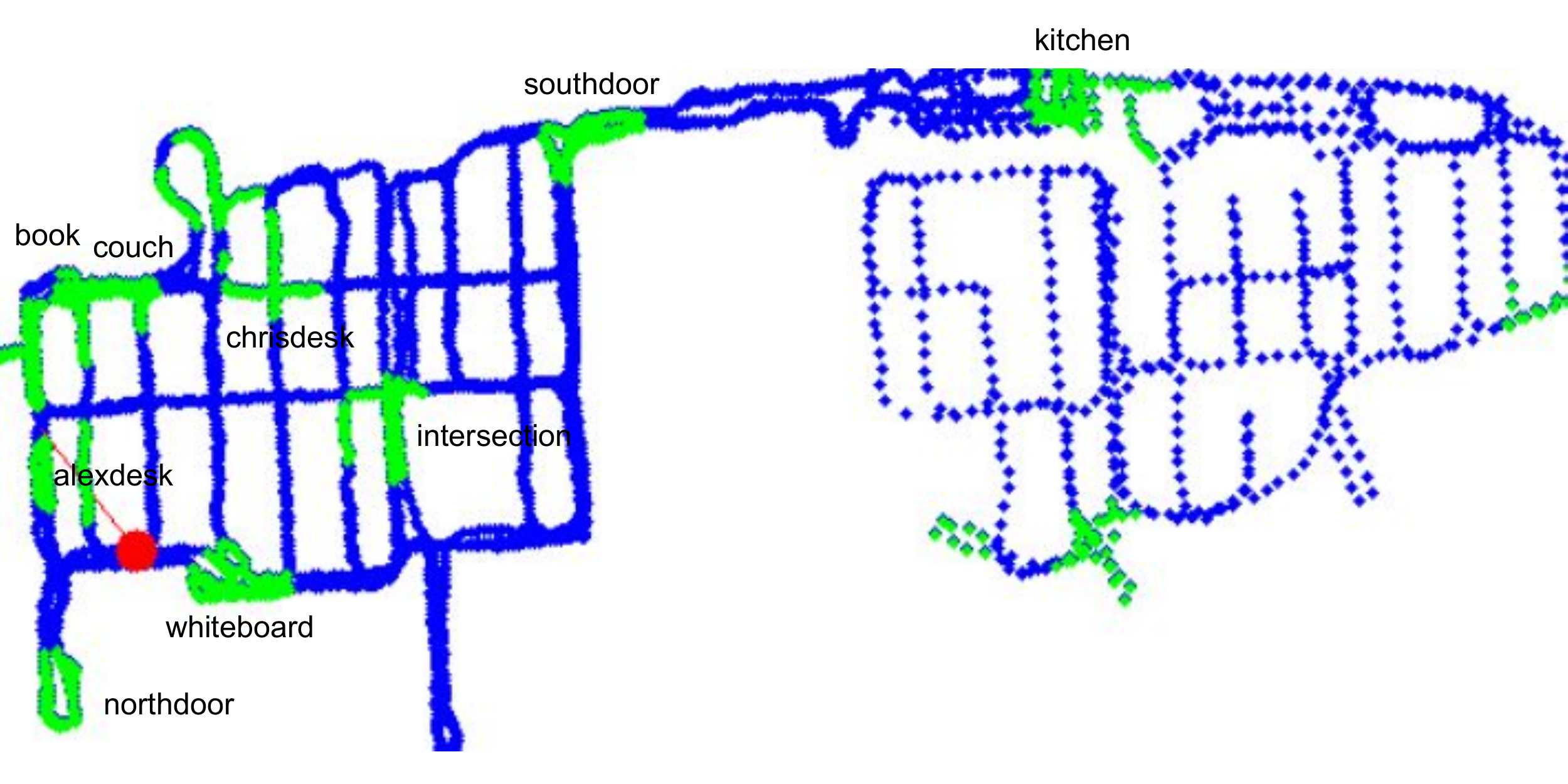}
\caption{\label{fig:targets}\env{} of the scanned building. Target, with 3m radius around them, marked in light green. Some parts of the building were scanned at different step size, as a way to demonstrate that such type of data can vary over space and time. The red dot is an example robot location and orientation.}
\end{figure}

For training, we scan a large $200m$ x $100m$ office building. In this scan, a panorama is captured every $0.5m$ during a traversal. Note that these scans were acquired months apart in time, and have differences in the height of the camera. For testing we have two setups: (i) a second scan of the same building (5835 distinct locations; see Fig.~\ref{fig:targets}). (ii) a real robot which runs the policy in the building.

\begin{figure*}
\centering
\includegraphics[width=0.9\textwidth]{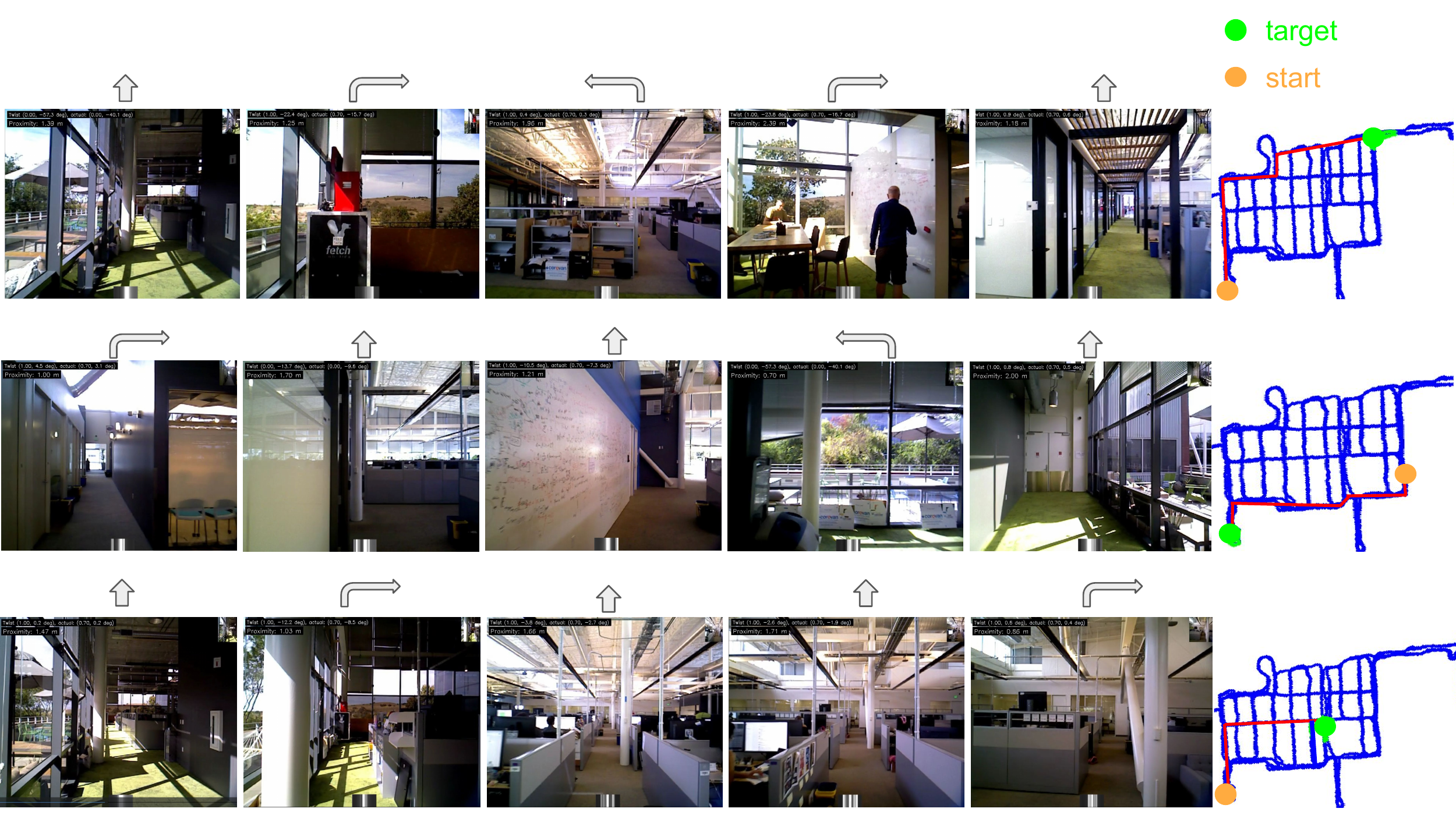}
\caption{\label{fig:real_robot_runs}Visualization of three randomly picked runs on the real robot, one run per row; each row showing representative frames for a decision the policy takes at junctions in the environment.}
\end{figure*}

\subsection{High Level Planning Policy}
We evaluate how well the high level policy performs in isolation of the low level policy. For this we use \env{}, where we train on data from $3$ traversals and test using data from a holdout traversal. This simulates the scenario where we navigate the same environment but see different images from different poses. At test time, the policy traverses the test \env{} scan and evaluates navigating to $10$ locations (see Fig.~\ref{fig:targets}). For each location, we randomly pick $20$ starting positions with each target location at least $30m$ away, resulting in $200$ runs.

The high level policy achieves near perfect success rate of $0.96$ on the test traversal. This performance is consistently high across all targets (at least $0.90$ success rate). The average run length is $78$ steps which corresponds to approx.~$70m$, resulting in $2.8km$ total.

\textbf{Generalization over Time} It is to be noted that $2/3$ of the training traversal was taken 4 months apart from the testing one. Being an office building, some areas of the environment have undergone changes (moved furniture, different furniture). This creates a realistic setup, but also a challenging one. 

Most of the failures were due to such changes. Also, some of the test traversal was missing areas present in the train one, which was causing additional failures, as the robot was seeking to go through these areas. Despite these challenges, our approach resulted in navigation performance close to perfect.

\begin{figure}
\centering
\includegraphics[width=0.75\columnwidth]{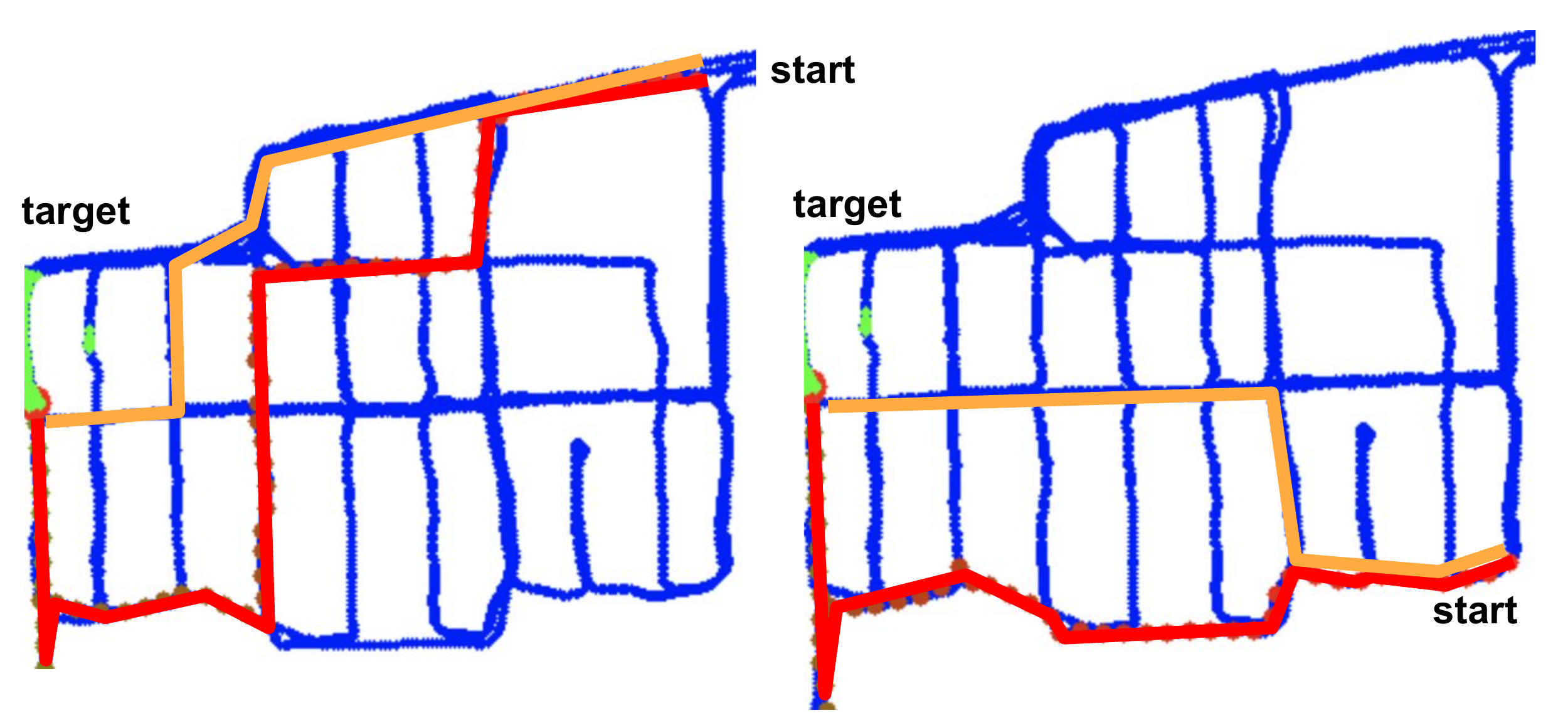}
\caption{\label{fig:diff-paths}Two runs in which the policy (in red) chose a different path than the optimal one (in orange)}
\end{figure}

The results also indicate that the policy generalizes to novel views and changes in the environment. The test traversal does not follow the train one and was recorded a month later at a different camera height. The policy does not perfectly follow the supervision (Fig.~\ref{fig:diff-paths}, right) and can recover from missed turns (Fig.~\ref{fig:diff-paths}, left). In aggregate, the trained policy learns to get to the target efficiently with an average run length $1.24$ times the shortest possible one.

\textbf{Generalization over Space} The training data traverses every area three times. Although the operator didn't intend to make the traversals different, they might cover the areas slightly differently. We train \name{} with 1, 2, and 3 traversals. The success rate for training with 2 and 3 traversals is $0.96$, while with only 1 traversal, the success rate drops to $0.75$. Some of the performance drop can also be attributed to the single traversal experiment using a scan captured at a higher viewpoint. This shows that more data from the scanned environment helps, but the optimal performance is reached without the need to re-scan the environment many times.

\textbf{Image-based Target Specification} Furthermore, our method allows more flexibility in specifying targets for the policy. While traditional navigation systems take a global target pose as an input, our training setup can allow arbitrary target images. To verify this, we run an experiment where instead of specifying the target as a one-hot vector, we provide an image from a target area. We input to the policy a target image sampled randomly from a target area. Due to the $24$ rotations at each location, there are usually several hundred views at each target location at each of the $77$ locations. At test time, we sample a target image in the same way. 

\begin{figure}
\centering
\includegraphics[width=0.75\columnwidth]{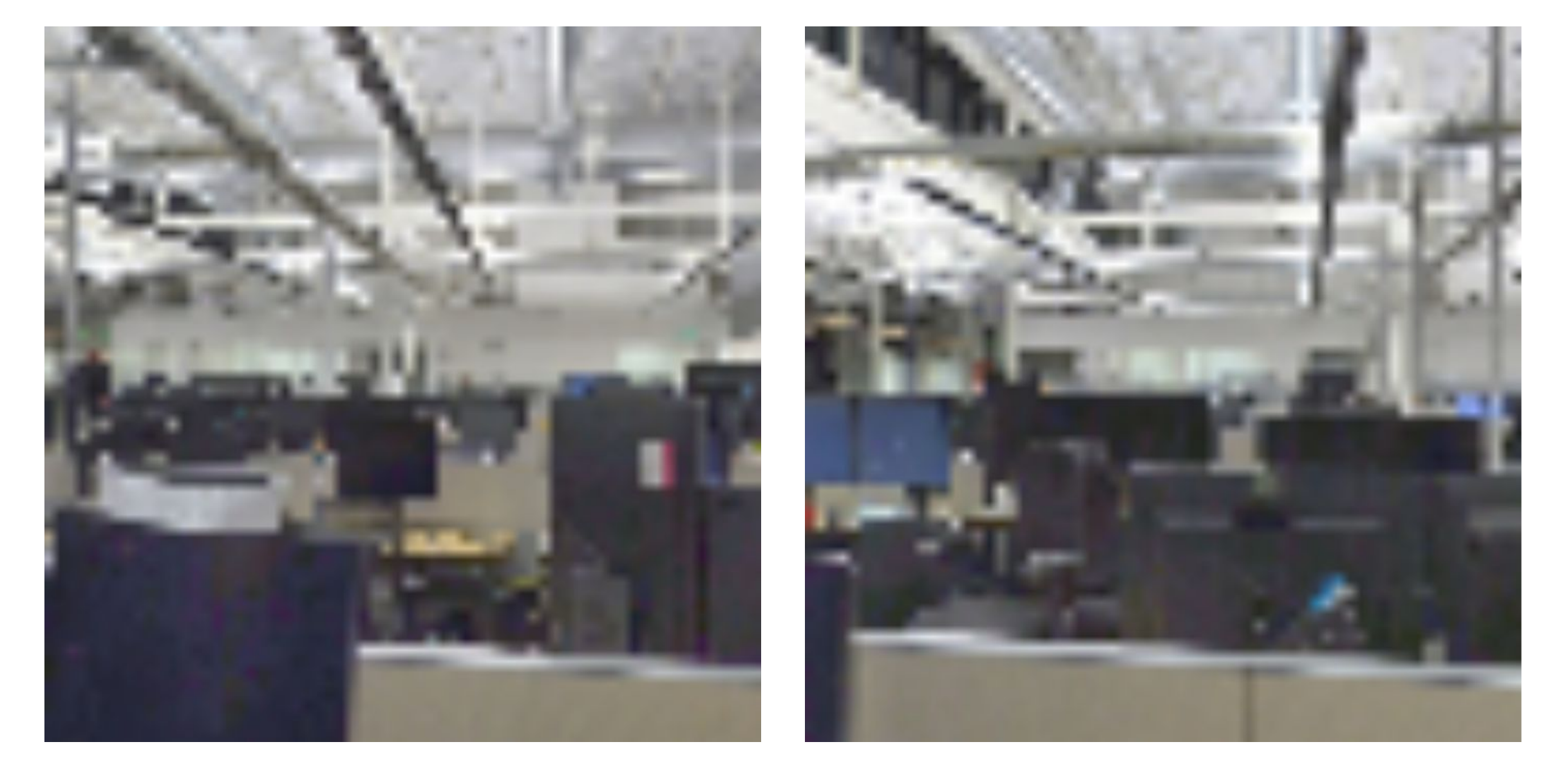}
\caption{\label{fig:ambiguous-target}Two different locations in the environment, showing the visual similarity of different possible targets.}
\end{figure}

Using this setup, we achieve $0.85$ success rate. Our analysis shows that the decrease in performance is mostly from receiving ambiguous target images. For example, an image of a wall or window provides very little information for the policy to determine where to go, or an image of a cubicle area may be hard to differentiate since there are a large number of cubicles in the environment (Fig.~\ref{fig:ambiguous-target}). To verify this, we also run an experiment where we also provide an additional target image at the same location but facing in the opposite direction to cover a larger field of view. With this additional input, we achieve $0.89$ success rate, showing that more views helps the policy disambiguate the target. These experiments also show that we can generalize to a large number of targets.

\textbf{Shortest Path Planner vs Human Demonstration} In Sec.~\ref{sec:high_environment} we define the loss using a set of navigation paths. In the experiments above, these are generated by a shortest path planner. In addition, we can use demonstrations of human navigation, which allows us to learn behaviors otherwise hard or impossible to encode in a traditional planner.

\begin{figure}
\centering
\includegraphics[width=0.75\columnwidth]{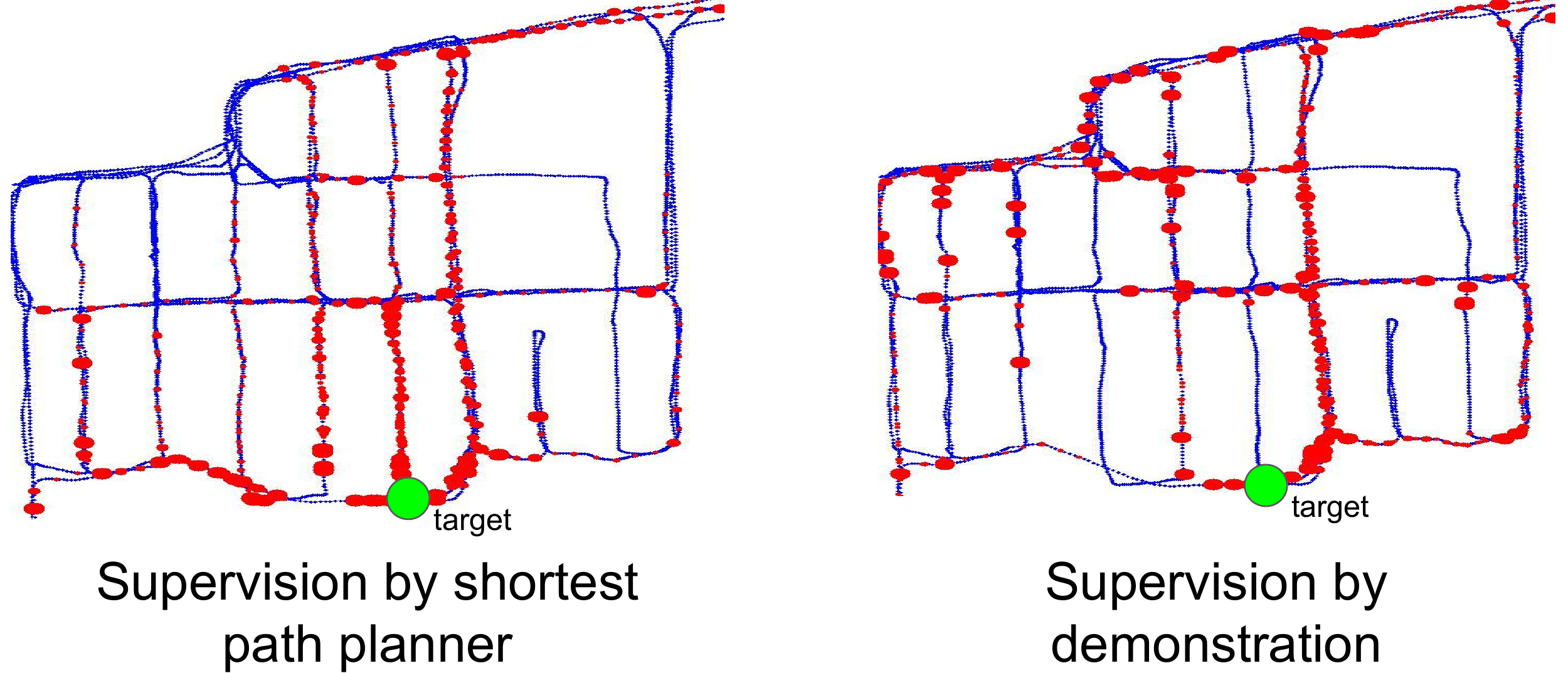}
\caption{\label{fig:shortest_path_vs_demos}We overlay in red a set of 50 test paths taken by the trained policy. The size of the red points demonstrates the visitation frequency across these paths.}
\end{figure}

For one of the targets, we ask an inhabitant of the office space to label plausible navigation trajectories leading to this target starting from various points in the building. We label $70$ demonstrations. A qualitative comparison between the shortest path supervision and demonstration supervision paths in the test \env{} is presented in Fig.~\ref{fig:shortest_path_vs_demos}. We see that the demonstration-based  navigation largely avoids the middle vertical sections. These correspond to paths through cubicle spaces, which are naturally avoided by humans as they are socially disturbing. This shows that a learning-based policy can learn socially aware behavior without manually encoding rules in the planner.

\subsection{Real Robot Experiment}
We use the Fetch robot mobile base~\cite{fetch_robot} to run real world experiments. For each of the 10 targets we perform 4 runs. We run the robot without any manual intervention. $1/3$ of the training scan was collected 7 months before the experiment, the rest $2/3$ was collected 3 months before.

\name{} achieves success rate of $0.80$ over these 40 runs. This shows that the policy is capable of learning long range planning while avoiding obstacles. As shown in Fig.~\ref{fig:real_robot_runs} these runs are long, over highly repetitive areas. The failed runs fall into roughly three categories: leaving the area it is trained for (4 runs); lost localization leading to spinning (1 run); collisions with objects due to low-level policy failures (2 runs).

To put these results into perspective, we evaluate an established off-the shelf SLAM-based model -- the ROS Navigation Stack~\cite{navstack}. This system requires scanning the environment to build a 2D occupancy map. \name{} only expects a collection of images with their relationship, without needing precise pose and mapping of surroundings. Furthermore, the ROS system requires manual localization at the beginning of every run, which we do not need. Finally, it utilizes more sensors -- 1D LiDAR and an RGBD head camera. \name{} requires a monocular RGB camera for the high-level policy, and 1D LiDAR for the low level. The ROS Navigation Stack also requires careful tuning -- manual cleaning of the occupancy map, manual tuning of hyperparameters for obstacle detection, etc. 

With this setup, it achieves $0.75$ success rate in reaching the $10$ targets. This shows that a learned system can outperform an established navigation system, even at a lower setup cost and manual work. The ROS system exhibits two major failures -- localization failure and collisions. \name{}, on the contrary, learns to extract from the RGB sensor the relevant information for best localization in the current environment, while ROS relies on non-adapted feature matching. Furthermore, the occupancy map used by ROS leads either to collisions or being stuck, depending on how the hyperparameters governing the safety margins are set. The learned policy seems to learn a better behavior. 
 


\begin{figure}
\centering
\includegraphics[width=0.90\columnwidth]{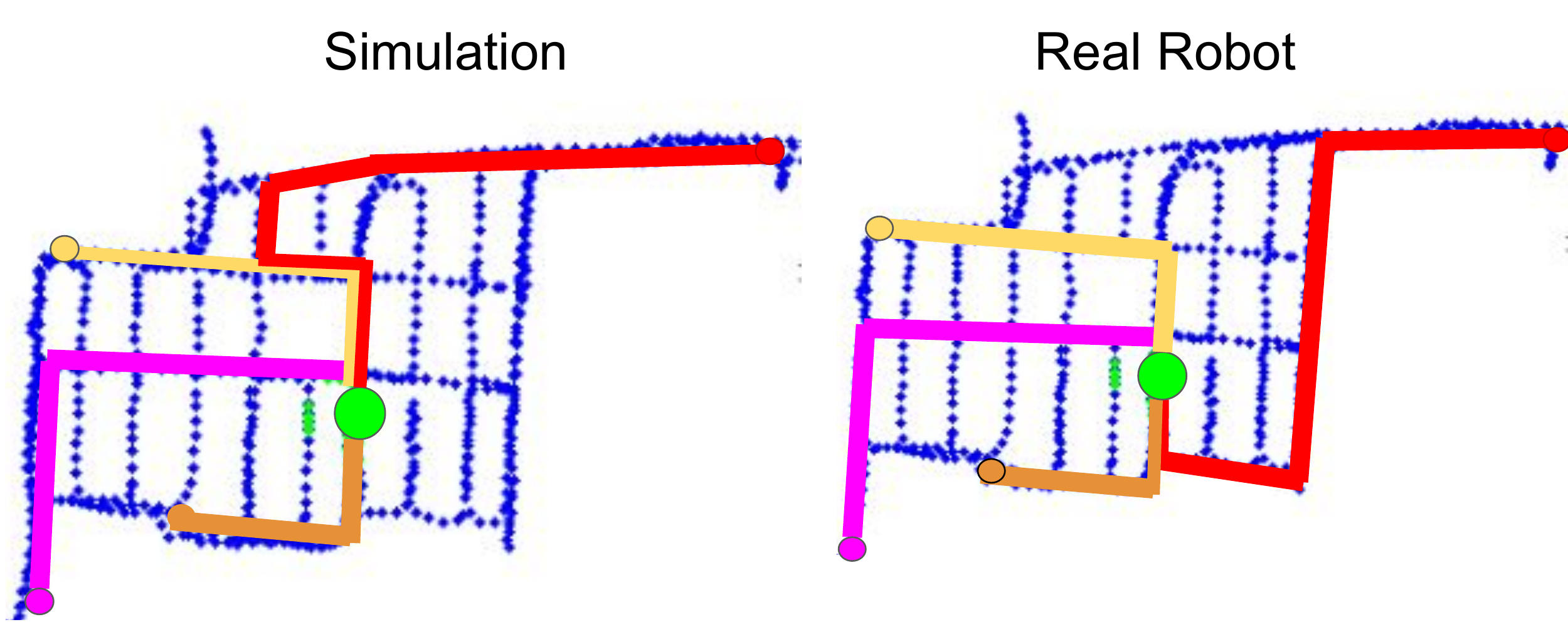}
\caption{\label{fig:sim-vs-real}Four runs in simulation (left) and in the real (right). The target is in green.}
\end{figure}

\textbf{\env{} vs Real Robot Runs} Another indication that the learned policy generalizes to unseen views and thus is robust in real settings is a comparison of runs in \env{} with runs on the real robot. In Fig.~\ref{fig:sim-vs-real} we show 4 runs with same start and target executed in simulation and in real. In 3 out of 4 cases the robot took identical paths. In one run (in red) it made a different but meaningful turn at the first intersection and then successfully reached the target in similar distance.

\subsection{Low Level Policy}

\textbf{Learned vs ROS} To investigate the importance of our learned low level policy, we experiment using the ROS local planner to execute high level commands from \name{}. This requires a pre-generated occupancy map, which our low level policy doesn't need. To determine whether we can step forward at any location, we use the local costmap, which fuses sensor data from a LiDAR scan with a pre-computed SLAM map, to determine if the area in front of the robot contains traversable space. We provide the planner with one of three relative poses according to which action the high level planner chose. We tune the local planner to allow a $0.5m$ radius tolerance for “forward” action and $0.1$ radians for “rotate” actions. These parameters and several others were chosen after 1 day of manual tuning and testing. 

\begin{figure}
\centering
\includegraphics[width=0.8\columnwidth]{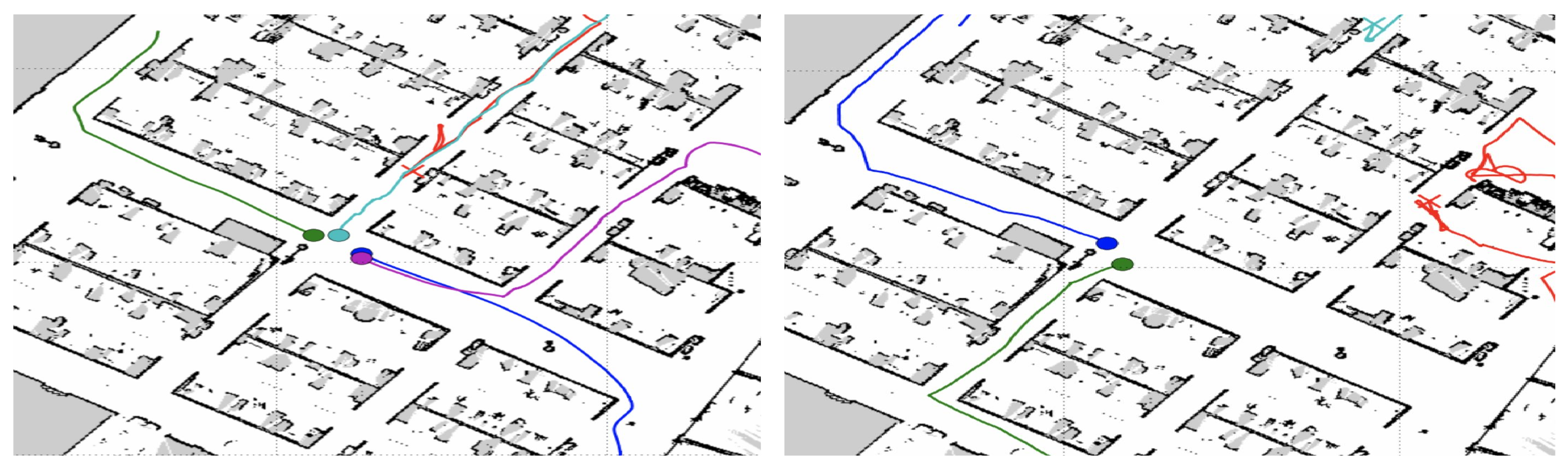}
\caption{\label{fig:navnet_vs_ros} Top-down view of runs. The end point is marked by a circle. Left: \name{}. Right: \name{} with low-level policy replaced by ROS Navigation stack.}
\end{figure}

We compare the two systems on 10 random runs. \name{} succeeds in 9, while \name{} with ROS succeeds in 6. Visualization of some of the successful trajectories are shown in Fig.~\ref{fig:navnet_vs_ros}. Failures roughly fall into the global policy taking the robot into areas where the local planner is not confident and gets stuck, unable to execute a “forward” command at any of the 24 rotations. More subjectively, the ROS local planner seems to result in more ``zig-zag`` paths while our policy maintains smoother and more central paths.

\begin{figure}
\centering
\includegraphics[width=0.75\columnwidth]{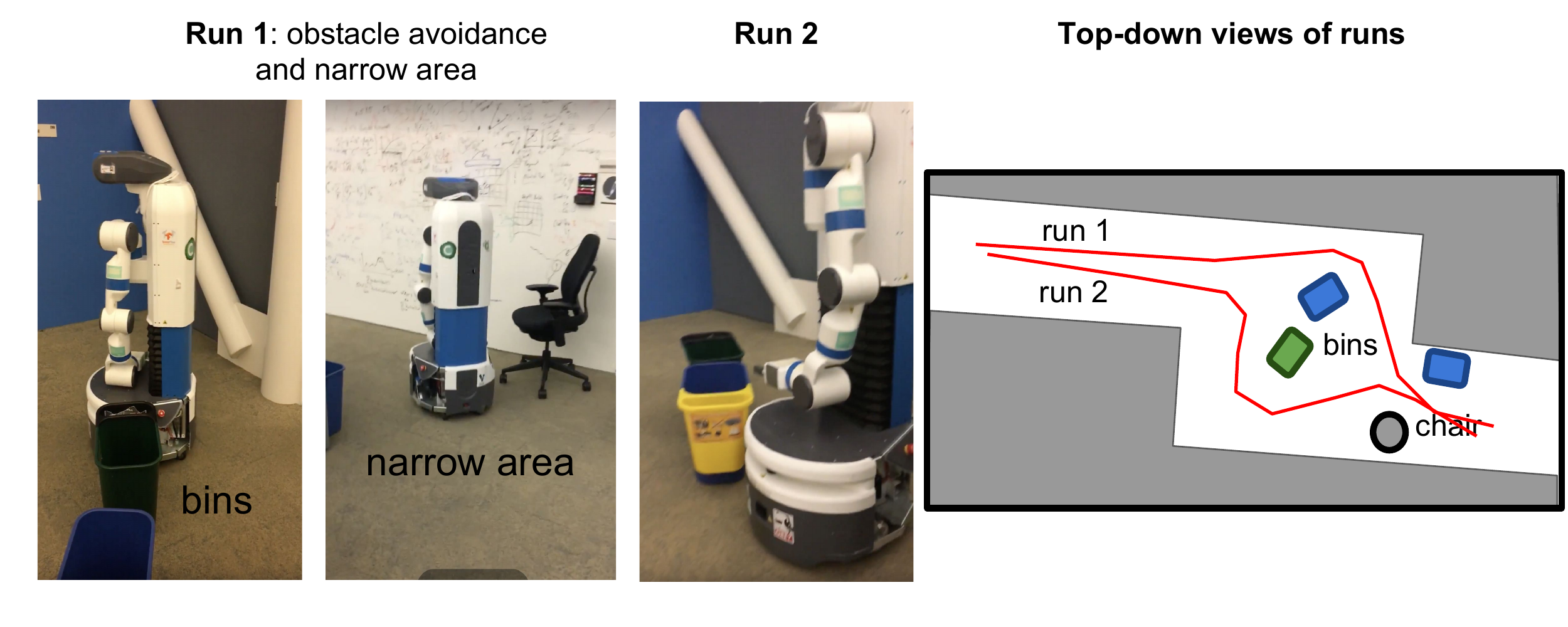}
\caption{\label{fig:lowlevel-example} Two runs with obstacles. Left: second person view while navigating around obstacles. Right: trajectories of the two runs.}
\end{figure}

\textbf{Qualitative Evaluation} We perform 4 runs on the robot with same start and target. For each of them, we place obstacles (2 bins and a chair) in the middle of the hallway for the robot to navigate through. In unobstructed settings, the robot goes through the middle of the hallway. With obstacles, the robot (i) still executes a correct high level plan and (ii) successfully changes the path execution to avoid the obstacles. As shown in Fig.~\ref{fig:lowlevel-example}, the area it has to go through is only approx.~$50\%$ wider than the robot base. The low level policy manages to get the robot safely around.

\section{Conclusion}\label{sec:conclusion}
In this paper we present a novel two level neural policy for navigation. We train it in a novel way using two different environments, which allows for easy deployment on a real robot. The system has high performance compared to SLAM-based methods and demonstrates that neural policies can be trained and deployed in the real world.

\section{Acknowledgement} The authors thank A.~Angelova, A.~Faust, J.~Kosecka for insightful comments; D.~Cheung, J.~Davidson, R.~Nguyen for organizational support.


\end{document}